\documentclass[runningheads]{llncs}

\usepackage{graphicx}%
\usepackage{multirow}%
\usepackage{amsmath,amssymb,amsfonts}%
\usepackage{mathrsfs}%
\usepackage{xcolor}%
\usepackage{textcomp}%
\usepackage{manyfoot}%
\usepackage{booktabs}%
\usepackage{algorithm}%
\usepackage{algorithmicx}%
\usepackage{algpseudocode}%
\usepackage{listings}%
\usepackage{float}
\usepackage{balance}
\usepackage{enumitem}
\usepackage{tabularx}
\usepackage{ragged2e}
\usepackage{booktabs}
\setlist{nosep}
\usepackage{wrapfig}
\usepackage[pagebackref,breaklinks,colorlinks,citecolor=blue]{hyperref}
\usepackage{orcidlink}

\usepackage{pifont}

\usepackage[table]{xcolor}
\usepackage{colortbl}
\definecolor{lightgray}{gray}{0.9}

\usepackage[accsupp]{axessibility}

\newcolumntype{L}[1]{>{\raggedright\arraybackslash}p{#1}}
\newcolumntype{R}[1]{>{\raggedright\arraybackslash}p{#1}}

\newcommand{\repthanks}[1]{\textsuperscript{\ref{#1}}}
\makeatletter
\patchcmd{\maketitle}
{\def\thanks}
{\let\repthanks\repthanksunskip\def\thanks}
{}{}
\patchcmd{\@maketitle}
{\def\thanks}
{\let\repthanks\@gobble\def\thanks}
{}{}
\newcommand\repthanksunskip[1]{\unskip{}}
\makeatother

\makeatletter
\DeclareRobustCommand\onedot{\futurelet\@let@token\@onedot}
\def\@onedot{\ifx\@let@token.\else.\null\fi\xspace}

\begin{document}


\title{GarmentSketch: Large-scale Sketch-to-Fashion Benchmark}

\titlerunning{GarmentSketch: Large-scale Sketch-to-Fashion Benchmark}


\author{
Duong-Duy-Khang Bui\inst{1,2}\orcidlink{0009-0005-7791-266X} \and
Minh-Tan Pham\inst{1,2}\orcidlink{0009-0004-8560-0431} \and
Tam V. Nguyen\inst{3}\orcidlink{0000-0003-0236-7992} \and
Minh-Triet Tran\inst{1,2}\orcidlink{0000-0003-3046-3041} \and
Trung-Nghia Le\thanks{Corresponding author.}\inst{1,2}\orcidlink{0000-0002-7363-2610}}

\institute{University of Science, Ho Chi Minh, Vietnam \and
Vietnam National University, Ho Chi Minh, Vietnam \and University of Dayton, Ohio, United States \\
\email{\{24C15009,24C15037\}@student.hcmus.edu.vn}\\
\email{tamnguyen@udayton.edu}\\
\email{\{tmtriet,ltnghia\}@fit.hcmus.edu.vn}}

\authorrunning{D-.-D.-K. Bui et al.}

\maketitle

\begin{abstract}
Fashion sketching is a cornerstone of design workflows, allowing rapid visualization of creative concepts prior to physical prototyping. Yet, progress in sketch-based fashion image synthesis has been hindered by the absence of large-scale, high-quality paired resources. To bridge this gap, we present GarmentSketch, a novel dataset comprising 26,249 fashion sketches across 21 garment categories, each paired with detailed textual descriptions. Captions were produced through a multi-stage pipeline that integrates multiple multimodal large language models (MLLMs) with human-in-the-loop refinement, ensuring both semantic accuracy and descriptive richness. We benchmark GarmentSketch on state-of-the-art generative models, providing baseline performance for sketch-guided text-to-image generation. Our experiments reveal both the promise and the current limitations of existing methods. By offering a comprehensive and richly annotated resource, GarmentSketch establishes a foundation for advancing sketch understanding, fine-grained fashion image generation, and creative human–AI collaboration in design. The dataset will be available at: \url{https://khangbdd.github.io/garmentsketch}


\end{abstract}

\section{Introduction}
Sketches play a fundamental role in the fashion design process. They serve as the primary medium for rapidly exploring concepts, styles, and garment details before physical prototyping~\cite{nguyen-truc2025advancing}. They provide designers with the flexibility to iterate freely, capture creative ideas, and convey visual narratives that extend beyond text limitations. As a result, sketching is both an artistic practice and a critical step in the broader workflow of ideation, communication, and refinement within the fashion industry.

\begin{figure}[t!]
    \centering
    \includegraphics[trim= 0 370 0 0,clip, width=\textwidth]{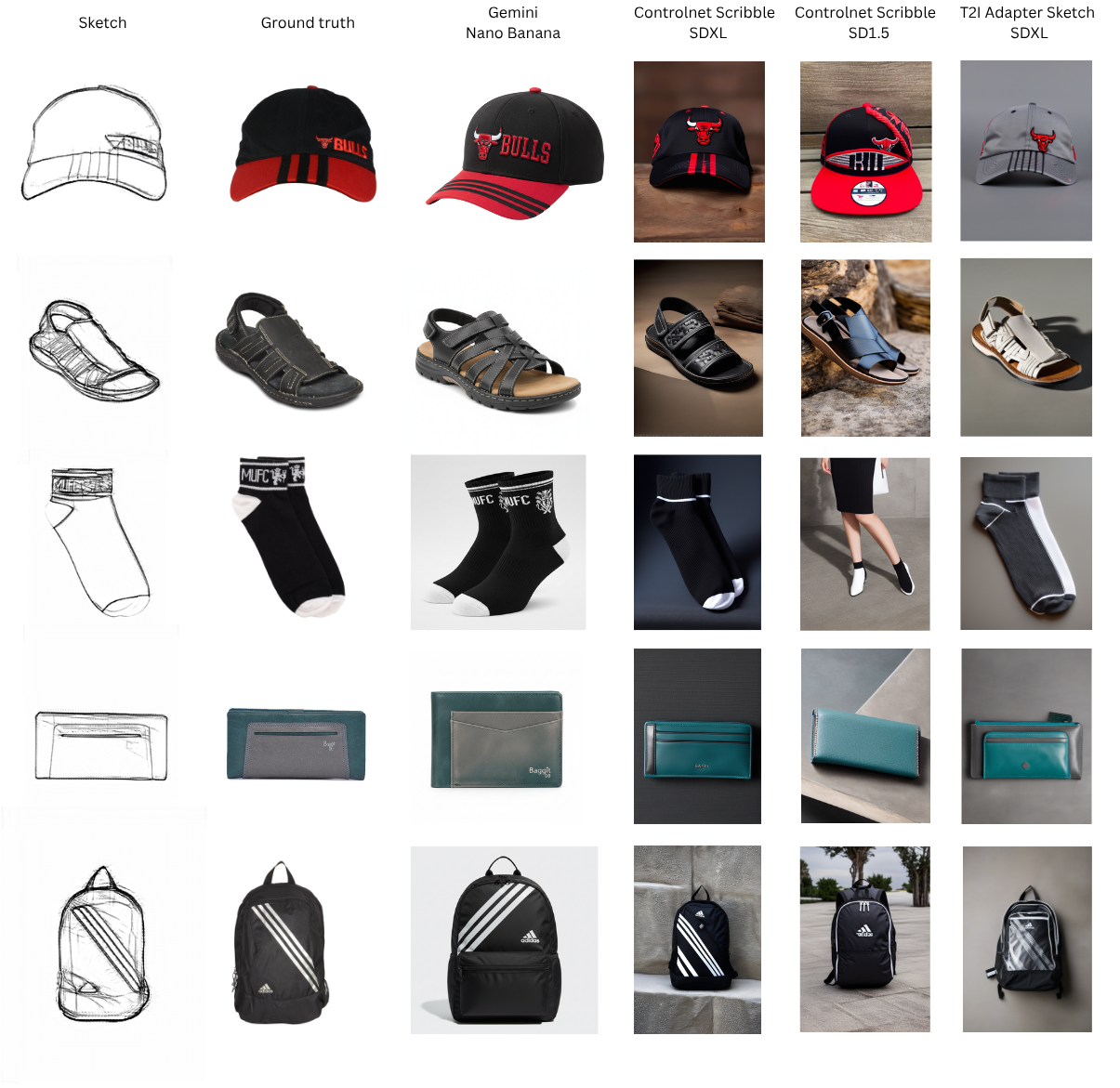}
    \vspace{-5mm}
    \caption{Limitations of existing methods. From left to right: input sketch, ground-truth, results from Controlnets\cite{ControlNetSD1.5}, T2I Adapter~\cite{t2i-adapter}, Gemini Nano Banana. The generated images often fail to preserve key design elements, losing fine details such as fabric draping, garment silhouette, and decorative embellishments.}
    \label{fig:limitations}
    \vspace{-5mm}
\end{figure}

Despite this importance, sketch-based image synthesis in the fashion domain remains significantly underexplored. Current generative models often struggle to interpret the abstract and highly stylized nature of sketches~\cite{ControlNetSD1.5}, which lack the rich pixel-level cues available in photographs. The challenge is further amplified by the scarcity of large-scale and high-quality paired resources that align sketches with corresponding images or descriptive annotations. Without such resources, models face difficulty in learning the complex mapping between sparse sketch lines and the rich textures, materials, and structures of garments~\cite{ControlNetSD1.5}. As a result, many existing systems fall short in producing outputs that faithfully capture the original intent of a designer~\cite{ControlNetSD1.5,t2i-adapter}. Generated images may lose critical details such as fabric draping, garment silhouette, or decorative elements (Fig.~\ref{fig:limitations}), thereby limiting their utility as reliable tools in the digital fashion design pipeline.

\begin{figure}[t!]
    \centering
    \includegraphics[width=\textwidth]{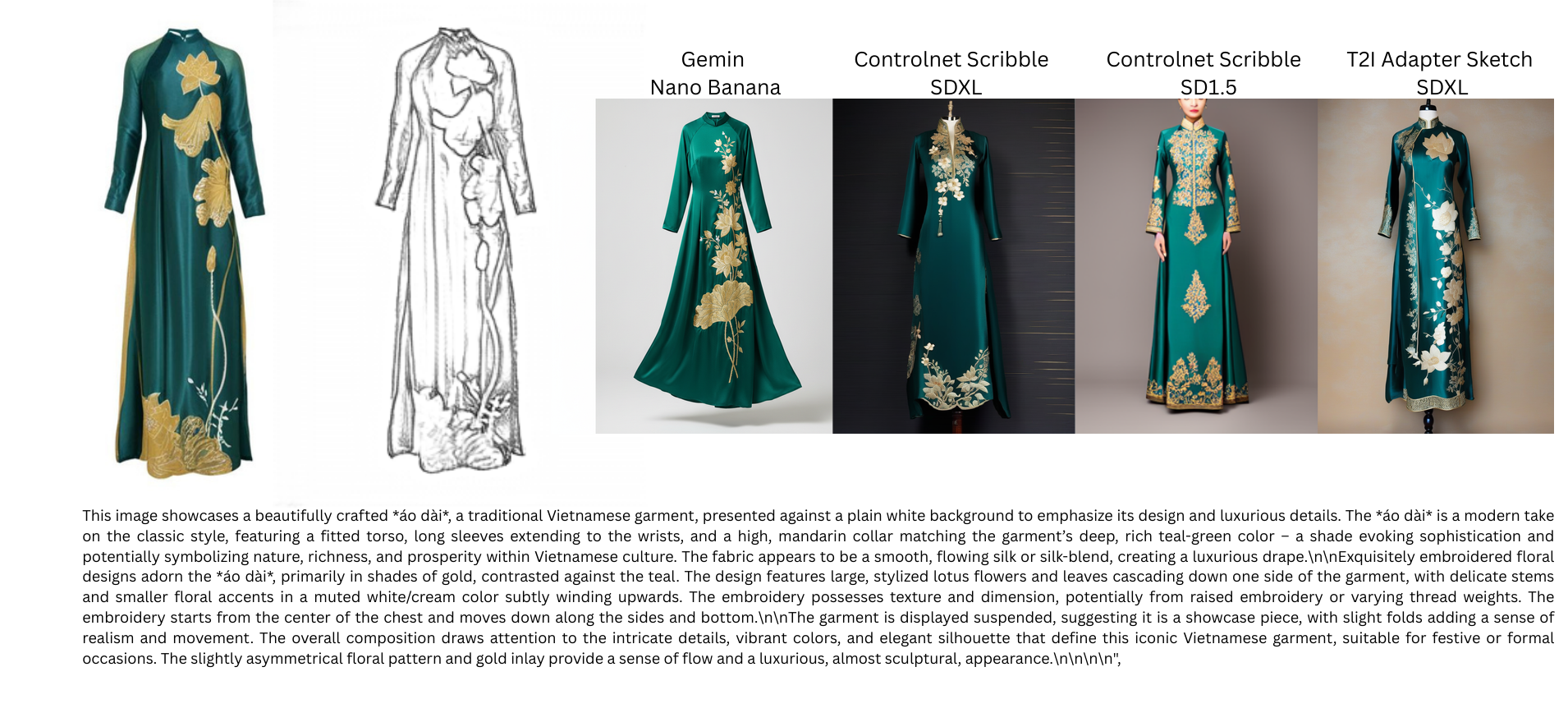}
    \vspace{-10mm}
    \caption{ A sample from the GarmentSketch dataset corresponds with the generated results from various benchmarking models. ({zoom to view the text clearer})}
    \label{fig:teaser}
    \vspace{-5mm}
\end{figure}

To address these challenges, we investigate the problem of sketch-guided text-to-image fashion generation. We introduce GarmentSketch, a large-scale dataset containing 26,249 unique fashion sketches spanning 21 garment categories. Each sketch is paired with a detailed descriptive caption generated through a multi-stage pipeline of multimodal large language models and subsequently refined through human verification to ensure semantic accuracy and contextual relevance. Fig.~\ref{fig:teaser} illustrates samples in our proposed GarmentSketch dataset.

{ Beyond introducing the dataset, we benchmark advanced diffusion and multimodal large language models on GarmentSketch to evaluate their preservation of design intent and garment details. Our central hypothesis is that aligning rich textual semantics with sparse sketches bridges the modality gap, enabling models to simultaneously preserve structural fidelity and synthesize complex details. Theoretically, GarmentSketch provides a foundational testbed for cross-modal representation learning, specifically aligning abstract structural priors with high-level semantics, advancing the understanding of controllable diffusion and multimodal generative AI. Ultimately, this curated resource aims to foster controllable models and support future research in fine-grained synthesis, sketch understanding, and human-AI collaboration.}

Our contributions are as follows:
\begin{itemize}
    \item We propose a data-curation pipeline that combines multiple LLMs with human-in-the-loop verification to produce high-quality, semantically rich descriptions for fashion sketches.
    \item We present GarmentSketch, a large-scale dataset of 26,249 sketch–caption pairs spanning 21 garment categories.
    \item We establish comprehensive benchmarks by evaluating several state-of-the-art multimodal image generation models, providing baseline performance and identifying key limitations.
\end{itemize}

\section{Related Work}\label{sec:related}

\subsection{Fashion Image Datasets}

{Large-scale datasets have heavily influenced fashion-related computer vision. DeepFashion~\cite{DeepFashion} provides extensive image annotations (categories, attributes, landmarks) for recognition, retrieval, and virtual try-on. FashionAI~\cite{fashionAI} and ModaNet~\cite{modanet} offer polygon-level annotations for clothing parsing and segmentation. Additionally, datasets like VITON~\cite{viton} provide paired garment–model images tailored for virtual try-on systems.}

{While these resources advance outfit manipulation tasks (e.g., attribute editing, virtual try-on, retrieval), they cannot adequately support design-oriented generation (Table~\ref{tab:dataset_comparison_final}). Crucially, they omit the large-scale pairings of sketch profiles, detailed textual descriptions, and corresponding images needed to study early-stage design workflows. Furthermore, existing sketch datasets fail to capture domain-specific fashion nuances; QuickDraw~\cite{quickdraw} offers massive volume but overly simplistic vector strokes, while Sketchy~\cite{sketchy} targets general object retrieval rather than fine-grained garment structures.}

{To bridge this fundamental divide, we introduce GarmentSketch, the first large-scale framework dedicated entirely to joint sketch-and-text-guided garment generation. Unlike prior works lacking this multimodal triad, GarmentSketch systematically integrates 26,249 informative fashion sketches across 21 garment categories, each paired with a reference image and a rich caption curated via an LLM-based pipeline and human verification. By supporting both structural adherence through sketches and semantic enrichment through language, our dataset provides a new foundation for research at the intersection of fashion design, multimodal learning, and generative modeling.}


\begin{table}[t!]
\centering
\caption{Comparison of existing fashion and sketch datasets with our GarmentSketch.}
\label{tab:dataset_comparison_final}
\begin{tabularx}{\textwidth}{@{} l | >{\RaggedRight}X | c | >{\RaggedRight}X | >{\RaggedRight}X @{}}
\toprule
\textbf{Dataset} & \textbf{Modality} & \textbf{Size} & \textbf{Primary Tasks} & \textbf{Description} \\
\midrule
DeepFashion~\cite{DeepFashion} 
& Images, attributes, landmarks 
& 800K images 
& Recognition, retrieval, virtual try-on 
& No sketches or captions \\
\midrule
FashionAI~\cite{fashionAI} 
& Images, hierarchical attributes, keypoints 
& 357K images 
& Attribute prediction, parsing 
& Limited design relevance \\
\midrule
ModaNet~\cite{modanet} 
& Images with polygon annotations 
& 55K images 
& Detection, segmentation 
& No sketches or captions \\
\midrule
VITON~\cite{viton} 
& Paired person and garment images 
& 16K pairs 
& Virtual try-on 
& Focused only on try-on task \\
\midrule
Sketchy~\cite{sketchy} 
& Sketches and images of general objects 
& 75K sketches 
& Sketch-based image retrieval 
& Not specialized for the fashion domain \\
\midrule
QuickDraw~\cite{quickdraw} 
& Sketches (Vector Strokes)
& 50M sketches 
& Sketch recognition 
& Too simplistic for garment design \\
\midrule
\rowcolor{lightgray} 
\textbf{GarmentSketch (Ours)} 
& Fashion sketches with rich captions 
& 26K sketches 
& Sketch \& text guided image generation
& First dataset for fashion sketches with rich captions \\
\bottomrule
\end{tabularx}
\vspace{-5mm}
\end{table}

\subsection{Image Generation}
Our research is positioned at the intersection of two major areas in generative
modeling: text-to-image synthesis and sketch-to-image translation. In this sec-
tion, we review key developments in both fields and highlight the gap that our
work aims to address.

\textbf{Text-to-Image. }
{Text-to-image generation has advanced rapidly, shifting from Generative Adversarial Networks (GANs) \cite{CycleGAN2017} to diffusion models \cite{SD,SDXL}. By guiding a denoising process via latent text prompts, diffusion models produce high-resolution, photorealistic images, revolutionizing content creation.}

\textbf{Sketch-to-Image. }
{Concurrently, sketch-to-image translation converts abstract line drawings into realistic images. Early methods approached this as a general image-to-image translation task using frameworks like Pix2Pix \cite{pix2pix} for paired data and CycleGAN \cite{CycleGAN2017} for unpaired datasets. Recent specialized models focus on the sparse, textureless nature of sketches to synthesize fine details like color and shading while respecting the input's structural core.}

\textbf{Sketch-Guided Text-to-Image. }
{Combining text and sketch guidance remains challenging. Text-to-image models~\cite{SD,SDXL} lack precise structural control, while sketch-to-image frameworks~\cite{self-conditional-GAN,quality-conditonal-GAN,democratisingsketch} preserve structure but lack semantic richness. Multimodal adapters like ControlNet~\cite{ControlNetSD1.5} and T2I-Adapter~\cite{t2i-adapter} combine both inputs but struggle to balance them, often losing fine details like fabric textures and silhouettes. Furthermore, because these models are trained on general-purpose datasets, they lack fashion domain knowledge and fail to capture subtle garment properties (e.g., draping, motifs) needed for professional workflows. To overcome the bottleneck, the absence of datasets aligning sketches, captions, and images, we introduce GarmentSketch as a foundation for balanced, fashion-aware multimodal generation.}

\textbf{Sketch-to-Fashion. }
{Most fashion generation research focuses on outfit manipulation tasks like virtual try-on~\cite{NguyenNgoc2023DMVTON} and virtual dressing\cite{shen2024IMAGDressing-v1}, leveraging paired datasets~\cite{DeepFashion,viton} for realistic bodily alignment. Other areas include attribute editing~\cite{fashionAI} and compatibility prediction~\cite{polyvore}, which uses specialized item sets to evaluate compatibility models~\cite{polyvore}.}

{In contrast, design-oriented generation from abstract sketches or concepts remains underexplored\cite{jin2024human}. This gap stems from three factors: first, existing fashion datasets lack sketch profiles or rich textual captions~\cite{DeepFashion,fashionAI,viton,modanet}. Second, generative models optimized for virtual try-on are strictly constrained to preserve the exact identity of existing garments, rendering them incapable of synthesizing novel, open-ended concepts for creative ideation~\cite{wang2025jco}. Finally, abstract, diverse sketches present severe challenges for models expecting pixel-level cues~\cite{ControlNetSD1.5,t2i-adapter}.}

{To address these limitations, we shift focus to design-oriented generation by introducing GarmentSketch. This large-scale dataset pairs sketches with detailed captions to enable systematic sketch-guided text-to-image synthesis. By combining structural sketch guidance with semantic caption enrichment, GarmentSketch helps models capture fine-grained creative details, bridging the gap between current generative methods and real-world fashion design workflows.}

\section{Proposed GarmentSketch Benchmark}

\textbf{Data Source. }
{We construct a diverse and globally balanced dataset from three complementary sources. Most categories utilize the Fashion Product Images dataset~\cite{Fashion_product_images_dataset}, providing over 44,000 high-resolution images with short descriptions. To enrich the upperwear category, we incorporated 2,000 random samples from the VITON-HD training set~\cite{VITON-HD}. Finally, to address cultural gaps in existing resources, we collected 650 online images of Eastern traditional costumes (e.g., the Vietnamese Ao Dai), ensuring a balanced representation of both Western and Eastern garments.}

\begin{figure}[t!]
    \centering
    \includegraphics[width=\textwidth]{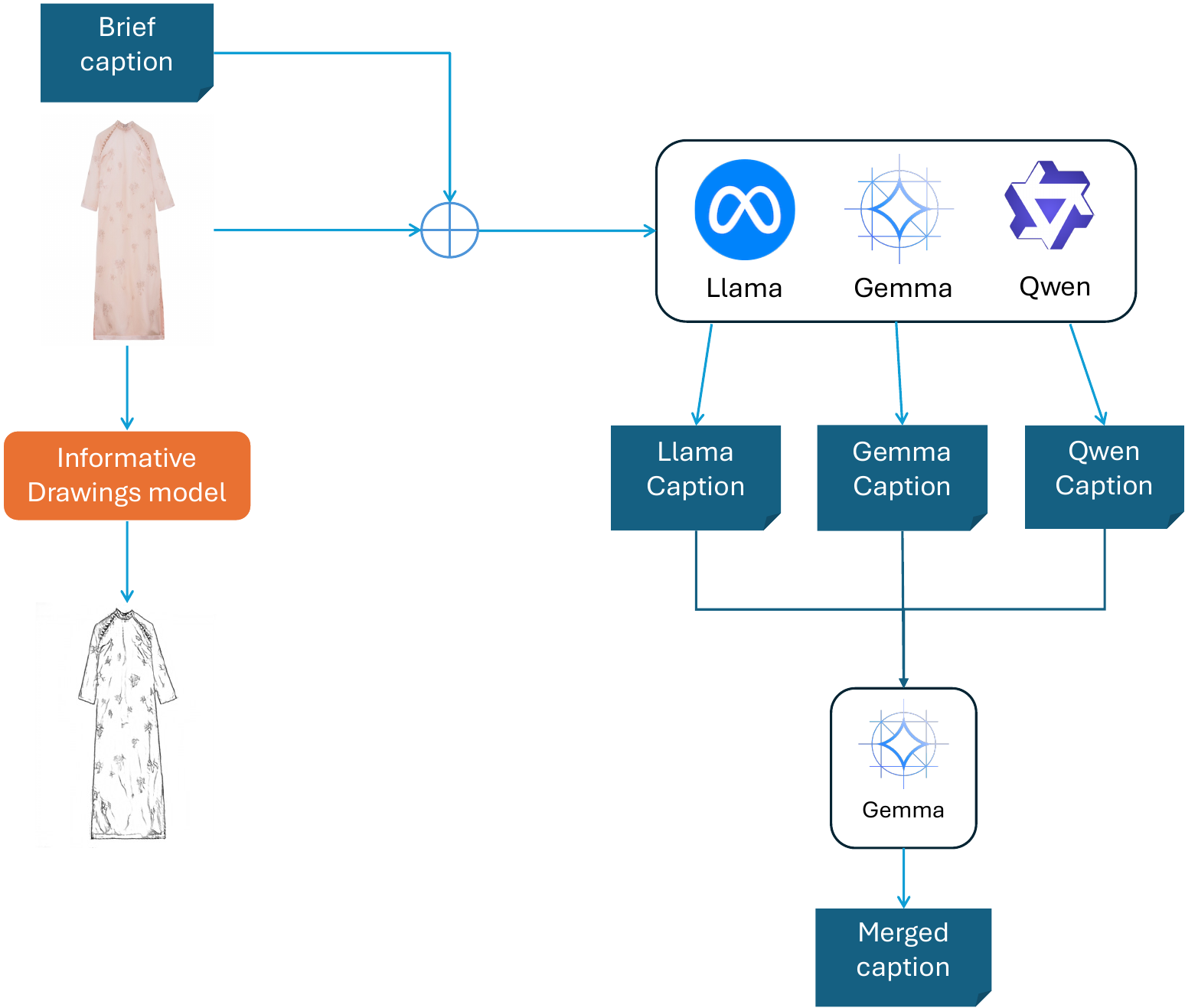}
    \vspace{-5mm}
    \caption{Overview of the data construction pipeline used to build the GarmentSketch dataset.}
    \label{fig:pipeline}
    \vspace{-5mm}
\end{figure}

\begin{figure}[t!]
    \centering
    \includegraphics[width=0.75\textwidth]{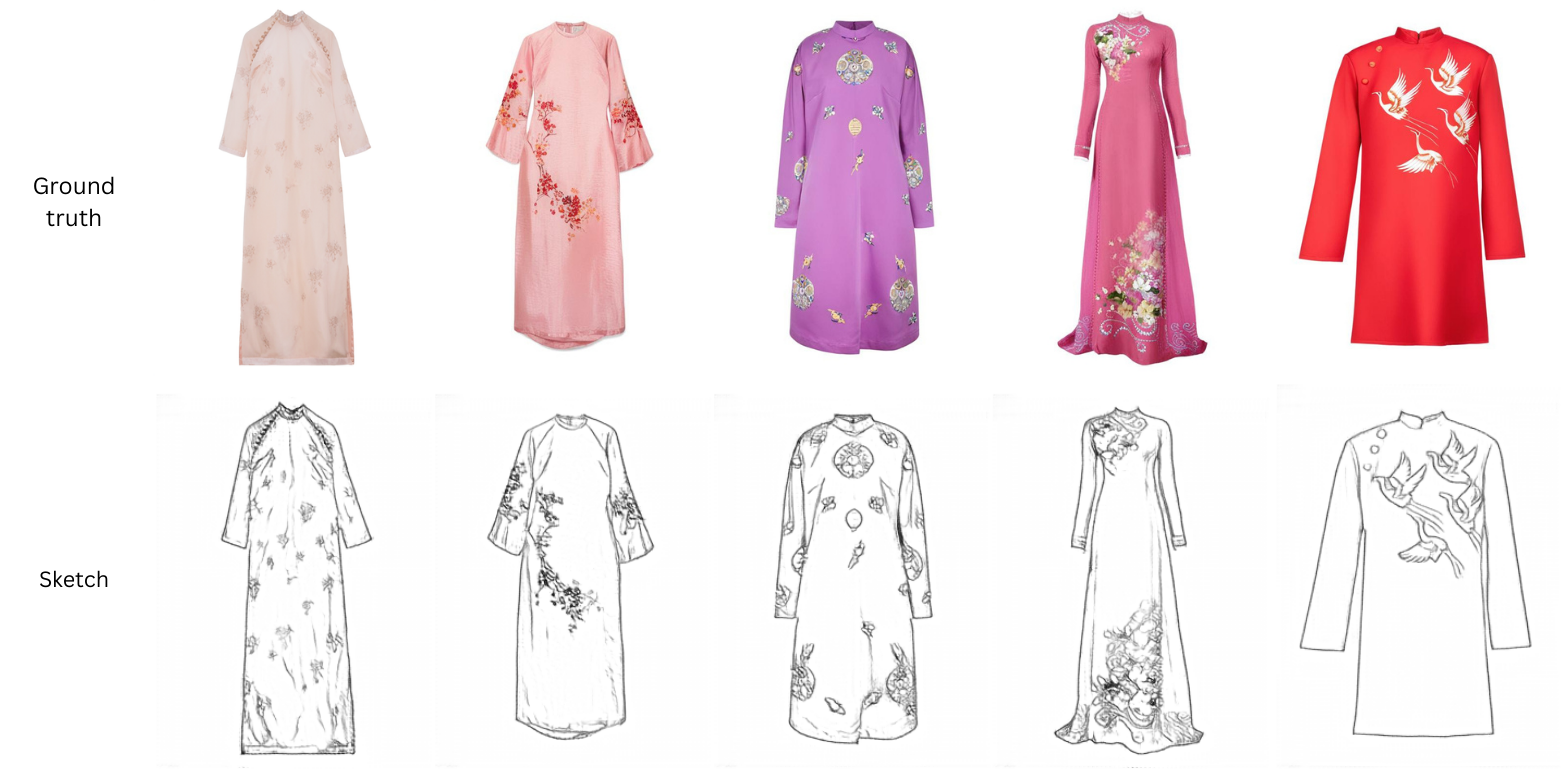}
    \includegraphics[width=0.75\textwidth]{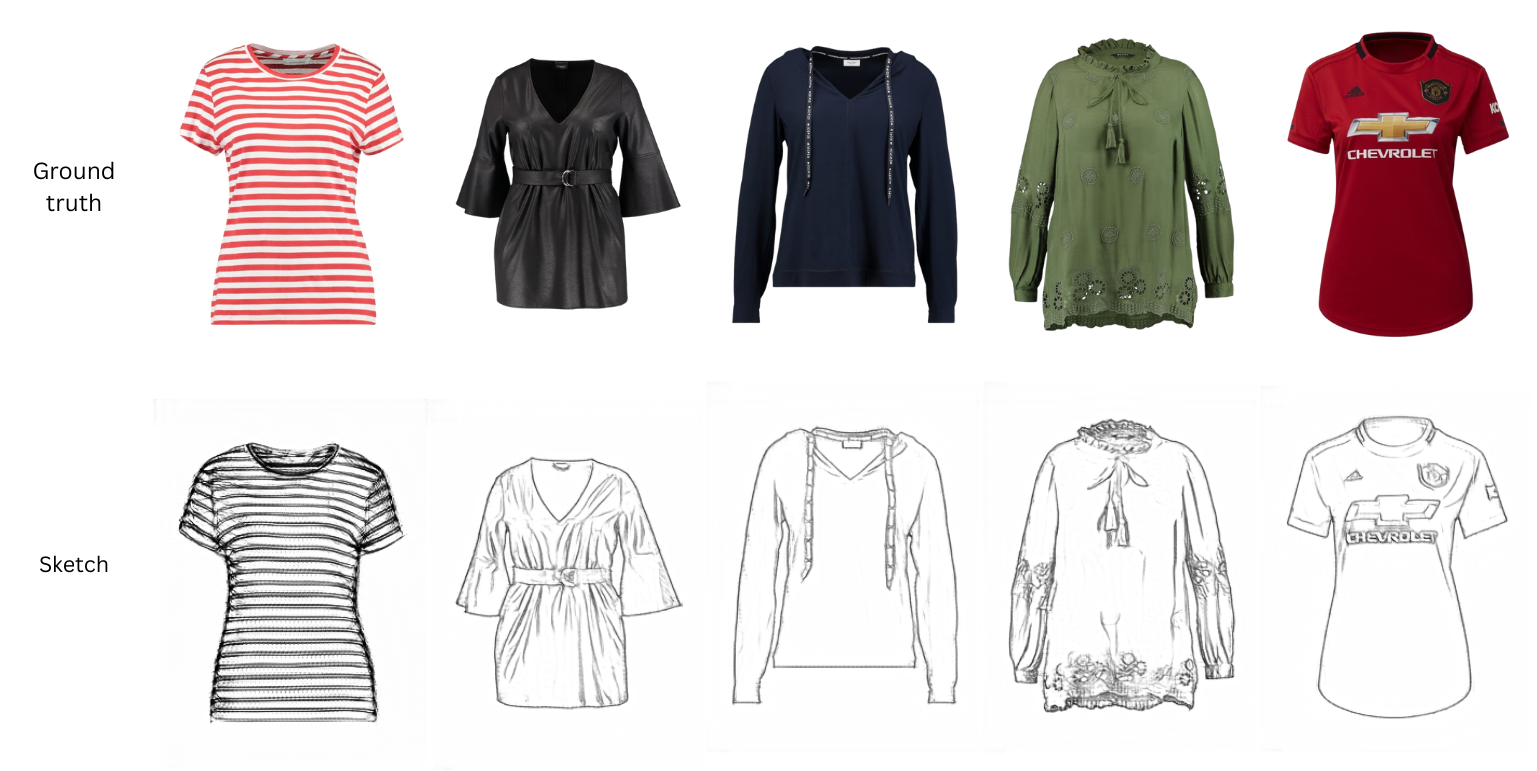}
    \includegraphics[width=0.75\textwidth]{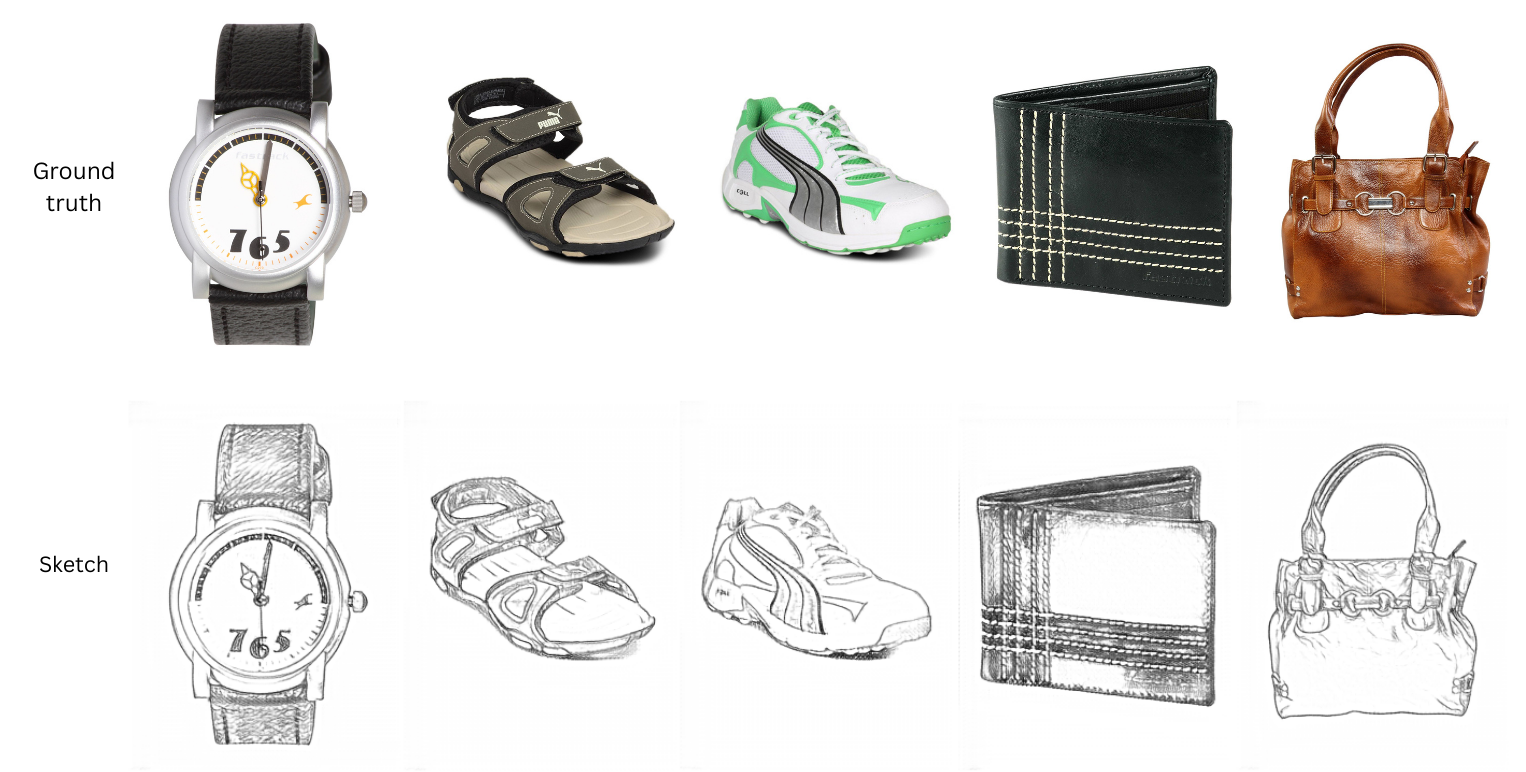}
    \vspace{-5mm}
    \caption{Examples of sketches generated by our proposed pipeline.}
    \label{fig:sketch}
    \vspace{-5mm}
\end{figure}

\begin{figure}[t!]
    \centering
    \includegraphics[trim={1cm 5cm 1cm 5cm}, clip, width=\linewidth]{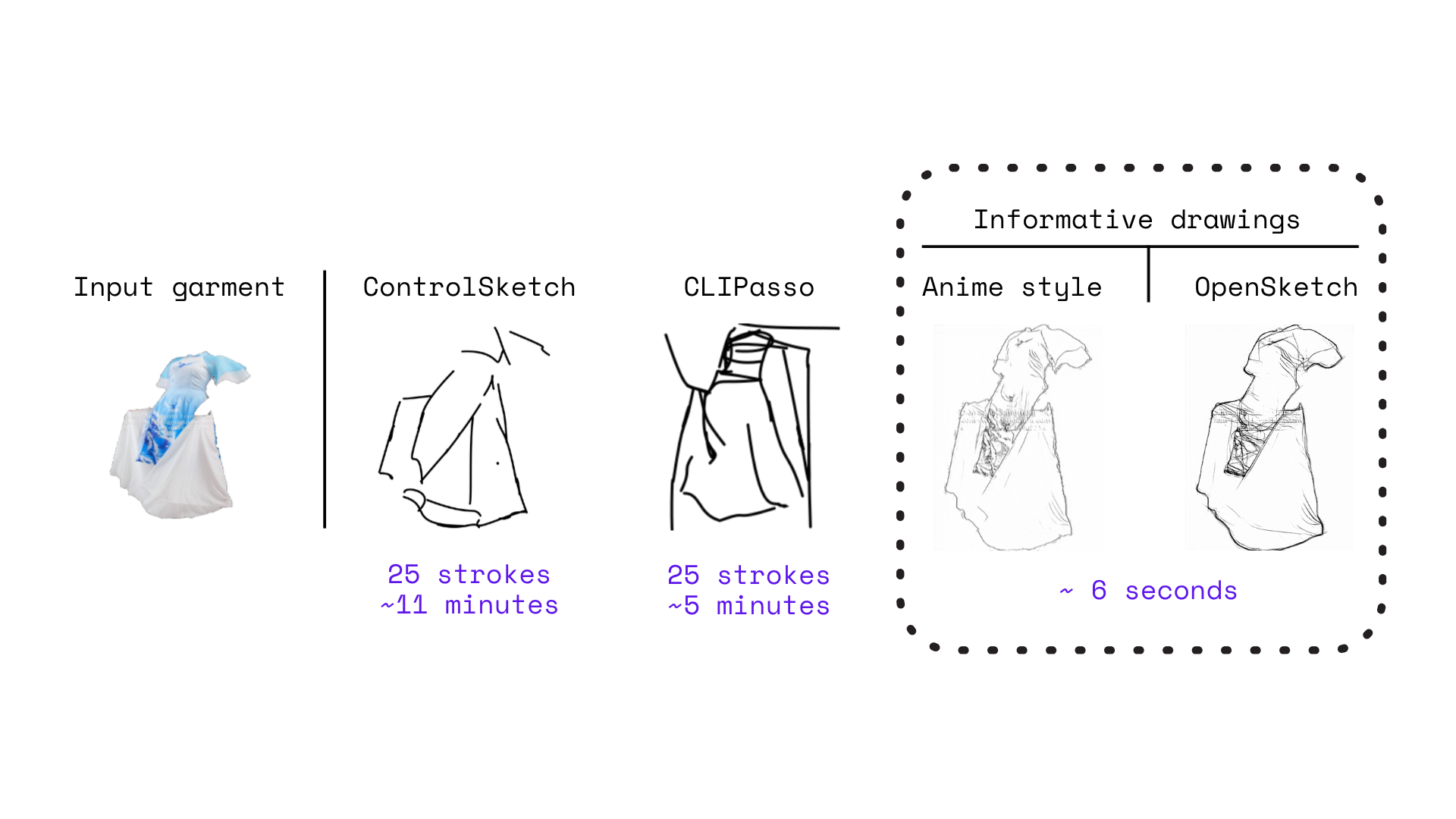}
    \caption{{Comparison of sketch generation methods.}}
    \label{fig:compare_sketch_creator}
\end{figure}

\textbf{Construction Pipeline. }
The construction of our dataset follows a dual-component pipeline, illustrated in Fig.~\ref{fig:pipeline}, consisting of two parallel workflows: (1) the generation of informative sketches from source images, and (2) the synthesis of rich, descriptive captions. On the left side of Fig.~\ref{fig:pipeline}, the sketch generation pipeline employs the anime-style Informative-Drawings model~\cite{InforDraw}, which converts photographs into line drawings. This approach is grounded in the observation that effective line drawings encode both shape and semantic meaning, serving as a compact yet informative representation of garment. {To justify our selection of the Informative-Drawings model, we evaluated it against other typical graphic tools used for sketch generation, such as ControlSketch~\cite{swiftsketch} and CLIPasso~\cite{clipasso} (illustrated in Figure \ref{fig:compare_sketch_creator}). Stroke-based optimization methods like ControlSketch~\cite{swiftsketch} and CLIPasso~\cite{clipasso} are computationally expensive, taking approximately 10 minutes to produce a 25-stroke output. Furthermore, these sparse vector strokes often fail to capture intricate garment patterns, textures, and delicate decorative elements. In contrast, the Informative-Drawings model is highly efficient, generating comprehensive sketches in approximately 6 seconds. Most importantly, it successfully captures the fine-grained design details of the apparel. Because it operates image-to-image rather than outputting mathematical vector strokes, it also better simulates real-world use cases where users capture and upload rasterized photographs of their physical hand sketches.} The resulting sketches accurately capture the geometric structure and salient semantic details of the original garment, providing reliable structural guidance for subsequent generative tasks. Figure~\ref{fig:sketch} illustrates examples of generated sketches.

On the right side of Fig.~\ref{fig:pipeline}, the caption generation pipeline is designed as a multi-stage process to produce high-quality, semantically rich descriptions. In the first stage, the source image and a preliminary description are processed by three multimodal large language models (MLLMs): Gemma-3-27b-it, Qwen2.5-VL-7B-Instruct, and Llama-4-Scout-17B-16E-Instruct, each generating diverse candidate descriptions. In the second stage, these outputs are consolidated using Gemma-3-27b-it, selected for its superior capacity for nuanced language understanding and compositional generation. The final caption represents a synthesized, polished description that integrates the complementary strengths of the individual models. To further ensure quality and consistency, captions are randomly sampled and verified by human annotators, aligning the descriptions closely with their corresponding images. 

\begin{figure}[t!]
    \centering
    \includegraphics[width=0.6\textwidth]{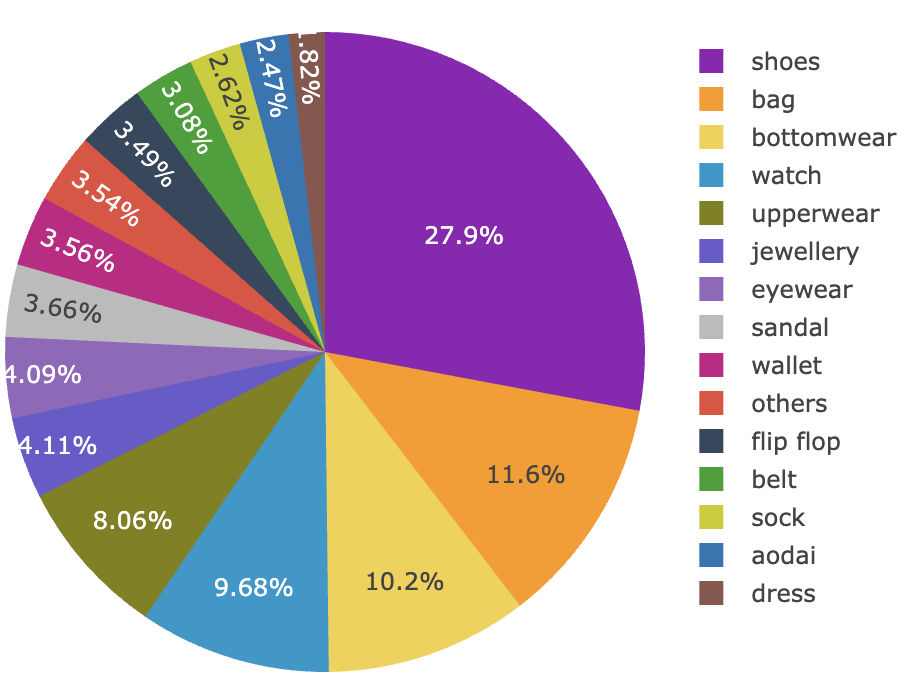}
    \caption{Distribution of garment category in GarmentSketch dataset. For visualization clarity, categories with fewer instances, such as Ties, Headwear, Scarves, Stoles, Mufflers, Cufflinks, and Gloves, are grouped under Others.}
    \label{fig:piechart}
\end{figure}

\textbf{Dataset Description. }
The proposed dataset contains 26,249 images organized into 21 fashion categories (Fig.~\ref{fig:piechart}). By balancing high-volume categories with culturally unique garments, our proposed GarmentSketch dataset offers both breadth and depth for advancing fashion generation research.

For experimental consistency, each category is pre-split into 70\% training and 30\% testing sets, ensuring standardized and reproducible evaluation across different studies.

{A deeper analysis of the category distribution (Fig.~\ref{fig:piechart}) reveals inherent data biases. First, a category imbalance heavily favors accessories: Shoes (27.9\%) and Bags (11.6\%) comprise nearly 40\% of the data, while core apparel like Upperwear (8.06\%) and Bottomwear (10.2\%) are less represented. This skew suggests that models will more readily learn the fixed structures of accessories than clothing. Second, a Western-centric cultural bias persists. Although we included 650 images of the Vietnamese Ao Dai for cultural inclusivity (approximately 2.5\% of the dataset), the vast majority of samples derive from Western e-commerce sources. Acknowledging these limitations is crucial, as they underscore the ongoing need for globally balanced data curation to prevent models from defaulting strictly to Western aesthetics.}

\section{Benchmarking}

\subsection{Evaluation Metrics}

We evaluate model performance on the proposed dataset using three quantitative metrics, following prior work on image generation~\cite{t2i-adapter,partialDiscrete,pictureThatSketch}. FID~\cite{fid} measures image quality and diversity, LPIPS~\cite{lpips} captures perceptual similarity to ground truth, and CLIPScore~\cite{clipscore} assesses semantic consistency with input text prompts.




\subsection{Benchmark Methods}

We evaluated four sketch-to-image models, including Gemini 2.5 Nano Banana, ControlNet Scribble SDXL, ControlNet Scribble SD1.5, and T2I-Adapter Sketch SDXL~\cite{t2i-adapter}, on the test set of GarmentSketch. All experiments were conducted in a zero-shot setting, without any fine-tuning, to assess generalization performance. Since many captions exceeded CLIP’s 77-token limit, we used the Gemma-3-27b-it model to generate shortened versions for the ControlNet and T2I-Adapter models.


\subsection{Qualitative Results}

\begin{figure}[t!]
    \centering
    \includegraphics[width=\textwidth]{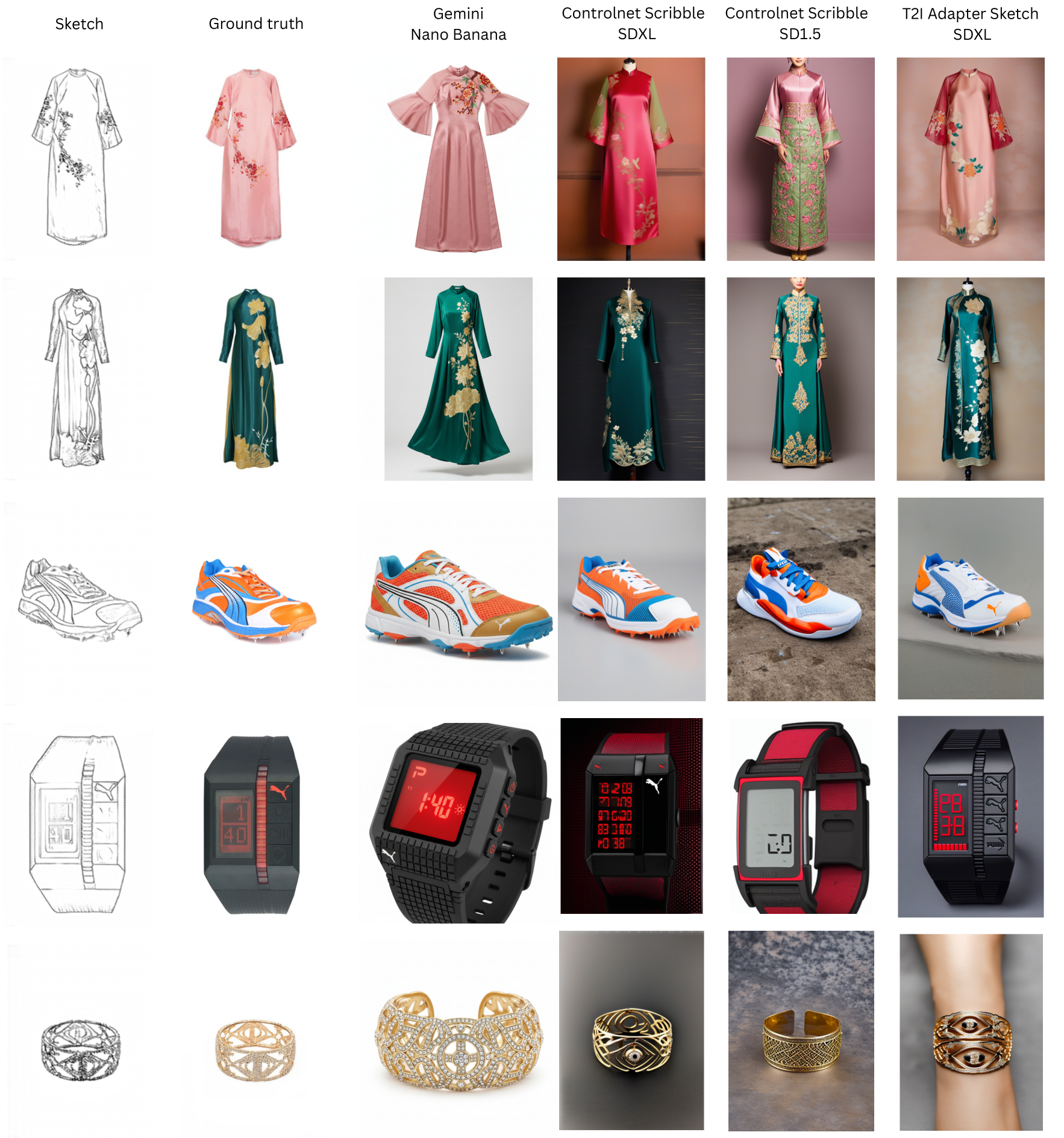}
    \vspace{-5mm}
    \caption{Qualitative evaluation of state-of-the-arts on GarmentSketch dataset.}
    \label{fig:Qualitative}
    \vspace{-5mm}
\end{figure}

Our qualitative evaluation reveals a clear trade-off between photorealism and structural fidelity across the benchmarked models. Gemini Nano Banana consistently produces the most visually appealing results, distinguished by high-quality textures and sophisticated lighting. Yet, this strength in photorealism often comes at the cost of faithfully preserving the input sketch, as the model frequently diverges from the intended silhouette and fine pattern details.

In contrast, ControlNet Scribble SDXL and T2I-Adapter Sketch SDXL exhibit stronger structural control, accurately reproducing garment shapes and forms as specified in the sketches. This makes them particularly suitable for design prototyping, where fidelity to form is critical. However, their outputs generally lack the rich textures and polished visual quality characteristic of Gemini Nano Banana’s generations.

The performance of ControlNet Scribble SD1.5 highlights the substantial progress enabled by newer architectures. It produced the least satisfactory results, often failing to preserve silhouettes and generating incoherent shapes. These shortcomings were especially evident with culturally specific garments such as the Vietnamese ao dai, which the model frequently misinterpreted, likely reflecting limited diversity in its training data. Its weaker performance establishes a baseline that underscores the improved fidelity and structural accuracy of the more recent SDXL-based ControlNet and T2I-Adapter models.


\subsection{Quantitative Results}

\begin{table}[t!]
\centering
\caption{Models comparison on the whole test set of the dataset.}
\begin{tabular}{l|ccc}
\toprule
\textbf{Model}            & \textbf{FID} $\downarrow$         & \textbf{LPIPS} $\downarrow$        & \textbf{CLIPscore} $\uparrow$   \\ \midrule
Gemini 2.5 Nano Banana                    & \textbf{17.50} & \textbf{0.40}  & 29.50 \\
ControlNet Scribble SDXL  & 29.18 & 0.60 & \textbf{30.10} \\
ControlNet Scribble SD1.5 & 36.08 & 0.70 & 29.02 \\
T2I adapter sketch SDXL   & 23.50 & 0.49 & 28.30 \\ \bottomrule
\end{tabular}
\label{tab:whole}
\vspace{-5mm}
\end{table}

As shown in Table~\ref{tab:whole}, Gemini Nano Banana delivers state-of-the-art performance on our benchmark. It achieves the best scores for both FID (17.50) and LPIPS (0.398), indicating superior visual fidelity, realism, and perceptual similarity to ground-truth images compared to the other models. ControlNet Scribble SDXL attains the highest CLIPScore (30.10), reflecting a slight advantage in semantic alignment with text prompts, though Gemini Nano Banana follows closely with a CLIPScore of 29.51. By contrast, ControlNet Scribble SD1.5 consistently underperforms across all three metrics, underscoring the progress achieved by more recent architectures. T2I-Adapter Sketch SDXL emerges as a competitive alternative, surpassing both ControlNet variants in FID and LPIPS, though it still falls short of Gemini Nano Banana in overall quality.

\begin{table}[t!]
\centering
\caption{FID comparison across categories for different models. CNS and T2IS stand for ControlNet Scribble and T2I Adapter Sketch, respectively.}
\begin{tabular}{lcccc}
\toprule
\textbf{Category} & \textbf{Gemini-2.5} & \textbf{CNS-SDXL} & \textbf{CNS-SD1.5} & \textbf{T2IS-SDXL} \\ \midrule
aodai       & \textbf{56.76} & 73.59 & 107.52 & 64.81 \\
bags        & \textbf{43.69} & 64.73 & 57.87  & 48.64 \\
belts       & 93.63          & 95.82 & 114.82 & \textbf{80.14} \\
bottomwear  & \textbf{30.76} & 70.49 & 62.76  & 58.63 \\
cufflinks   & \textbf{221.68} & 233.34 & 256.82 & 226.79 \\
dress       & \textbf{63.67} & 80.08 & 87.84  & 80.08 \\
eyewear     & \textbf{38.14} & 56.46 & 76.44  & 39.98 \\ 
flip flop   & \textbf{80.19}  & 104.34 & 125.56 & 87.49 \\
gloves      & \textbf{242.38} & 287.01 & 259.11 & 267.92 \\
headwear    & 154.85          & 181.62 & 227.39 & \textbf{153.06} \\
jewellery   & \textbf{94.51}  & 118.19 & 130.39 & 116.33 \\
mufflers    & \textbf{186.87} & 264.06 & 212.64 & 207.56 \\
sandals     & \textbf{50.04}  & 73.74  & 83.22  & 63.40 \\
scarves     & \textbf{134.25} & 174.04 & 202.78 & 152.05 \\ 
shoes      & \textbf{21.99}  & 30.68  & 45.83  & 26.56 \\
socks      & \textbf{82.06}  & 103.75 & 123.40 & 85.73 \\
stoles     & \textbf{124.63} & 152.66 & 167.87 & 156.04 \\
ties       & 144.53          & 206.33 & 200.02 & \textbf{131.00} \\
upperwear  & \textbf{35.60}  & 53.24  & 98.24  & 45.54 \\
wallets    & \textbf{73.34}  & 114.03 & 102.70 & 99.69 \\
watches    & \textbf{30.51}  & 34.57  & 41.54  & 32.24 \\ 
\rowcolor{lightgray} All & \textbf{17.50} & 29.18 & 36.08 & 23.50 \\
\bottomrule
\label{tab:fid}
\end{tabular}
\vspace{-5mm}
\end{table}

\begin{table}[t!]
\centering
\caption{LPIPS comparison across categories for different models. CNS and T2IS stand for ControlNet Scribble and T2I Adapter Sketch, respectively.}
\begin{tabular}{lcccc}
\toprule
\textbf{Category} & \textbf{Gemini-2.5} & \textbf{CNS-SDXL} & \textbf{CNS-SD1.5} & \textbf{T2IS-SDXL} \\ \midrule
aodai       & \textbf{0.3634} & 0.6415 & 0.6891 & 0.5557 \\
bags        & \textbf{0.4316} & 0.6268 & 0.6767 & 0.4642 \\
belts       & \textbf{0.3365} & 0.6186 & 0.7182 & 0.4556 \\
bottomwear  & \textbf{0.3956} & 0.5748 & 0.6274 & 0.4858 \\
cufflinks   & \textbf{0.3268} & 0.5241 & 0.7703 & 0.4943 \\
dress       & \textbf{0.3694} & 0.5619 & 0.6325 & 0.4862 \\
eyewear     & \textbf{0.3179} & 0.5410 & 0.7422 & 0.4237 \\ 
flip\_flop  & \textbf{0.3474} & 0.6883 & 0.8200 & 0.5266 \\
gloves      & 0.4958          & 0.5838 & 0.7620 & \textbf{0.4788} \\
headwear    & \textbf{0.4379} & 0.6862 & 0.7431 & 0.5408 \\
jewellery   & \textbf{0.3977} & 0.6425 & 0.7165 & 0.5468 \\
mufflers    & \textbf{0.4216} & 0.7963 & 0.7527 & 0.5896 \\
sandals     & \textbf{0.3734} & 0.6528 & 0.7634 & 0.5166 \\
scarves     & \textbf{0.4127} & 0.6802 & 0.7067 & 0.5168 \\ 
shoes      & \textbf{0.3961} & 0.5837 & 0.7539 & 0.4931 \\
socks      & \textbf{0.5104} & 0.6549 & 0.7398 & 0.5208 \\
stoles     & \textbf{0.4119} & 0.6284 & 0.6439 & 0.5455 \\
ties       & \textbf{0.3164} & 0.9032 & 0.8530 & 0.6030 \\
upperwear  & \textbf{0.4103} & 0.5949 & 0.6631 & 0.5242 \\
wallets    & \textbf{0.4753} & 0.6994 & 0.7723 & 0.5350 \\
watches    & \textbf{0.3931} & 0.5457 & 0.5704 & 0.4363 \\ 
\rowcolor{lightgray} All & \textbf{0.40}  & 0.60  & 0.70  & 0.49 \\
\bottomrule
\label{tab:lpips}
\end{tabular}
\vspace{-5mm}
\end{table}

\begin{table}[t!]
\centering
\caption{CLIPScore comparison across categories for different models. CNS and T2IS stand for ControlNet Scribble and T2I Adapter Sketch, respectively.}
\begin{tabular}{lcccc}
\toprule
\textbf{Category} & \textbf{Gemini-2.5} & \textbf{CNS-SDXL} & \textbf{CNS-SD1.5} & \textbf{T2IS-SDXL} \\ \midrule
aodai       & 28.83 & \textbf{30.40} & 29.22 & 27.63 \\
bags        & \textbf{26.82} & 26.39 & 25.40 & 23.70 \\
belts       & 31.33 & 32.44 & 28.50 & \textbf{33.13} \\
bottomwear  & 23.05 & \textbf{23.61} & 22.85 & 20.96 \\
cufflinks   & \textbf{46.80} & 46.64 & 46.08 & 50.88 \\
dress       & \textbf{31.53} & 28.10 & 27.70 & 31.13 \\
eyewear     & 36.26 & \textbf{38.69} & 34.93 & 36.18 \\ 
flip\_flop  & \textbf{30.28} & 28.73 & 28.07 & 29.00 \\
gloves      & 33.73 & \textbf{37.13} & 35.63 & 32.99 \\
headwear    & 28.87 & \textbf{30.65} & 29.39 & 30.04 \\
jewellery   & 30.33 & \textbf{31.77} & 29.86 & 28.00 \\
mufflers    & 17.90 & 19.11 & 17.16 & \textbf{20.56} \\
sandals     & 31.98 & 30.72 & 27.99 & \textbf{33.57} \\
scarves     & 33.62 & 33.76 & 32.72 & \textbf{34.52} \\ 
shoes      & 29.93 & \textbf{31.97} & 30.92 & 29.52 \\
socks      & 33.00 & \textbf{34.95} & 32.97 & 32.39 \\
stoles     & 23.12 & \textbf{24.79} & 23.37 & 23.16 \\
ties       & 26.57 & \textbf{27.51} & 25.51 & 26.37 \\
upperwear  & \textbf{22.26} & 21.53 & 22.24 & 20.61 \\
wallets    & \textbf{34.90} & 33.25 & 32.85 & 32.94 \\
watches    & \textbf{36.09} & 35.81 & 35.59 & 33.35 \\ 
\rowcolor{lightgray} All & 29.50          & \textbf{30.10} & 29.02 & 28.30 \\
\bottomrule
\label{tab:clipscore}
\end{tabular}
\vspace{-5mm}
\end{table}

Per-category results (Tables~\ref{tab:fid}, \ref{tab:lpips}, \ref{tab:clipscore}) show that Gemini Nano Banana dominates overall, achieving the best FID in 17 of 21 categories (e.g., shoes: 21.99, bottomwear: 30.76) and the best LPIPS in 20 of 21, with gloves as the only exception (0.4788 vs. 0.4958 for T2I-Adapter). T2I-Adapter Sketch SDXL demonstrates selective strengths, leading in FID for belts, headwear, and ties, and in LPIPS for gloves. For CLIPScore, performance is more balanced: ControlNet Scribble SDXL leads in 8 categories, Gemini Nano Banana in 7, and T2I-Adapter in 4 (e.g., scarves: 34.52). These numbers highlight Gemini Nano Banana’s consistent superiority, while ControlNet Scribble SDXL and T2I-Adapter offer complementary strengths in semantic alignment and structured garments. These results underscore Gemini Nano Banana’s reliability for high-quality image generation, while highlighting the complementary advantages of ControlNet and T2I-Adapter in semantic alignment and specialized garment structures.


\section{Discussions}

{Our benchmark reveals a fundamental trade-off between photorealism and structural fidelity. Gemini Nano Banana produces highly realistic, visually appealing images but often deviates from the input sketch's precise structure. Conversely, ControlNet and T2I-Adapter excel at preserving garment shapes but lack comparable photorealism. Furthermore, ControlNet Scribble SDXL shows an edge in semantic alignment, while T2I-Adapter demonstrates robustness for structured accessories like belts and gloves. These findings emphasize the need for future models that can balance creative flexibility with structural control.}

In addition, inconsistent performance on culturally specific garments, such as the Ao Dai, underscores a data bias in current foundational models. While our GarmentSketch dataset takes an initial step toward addressing this limitation, advancing the field will require more diverse datasets that capture global cultural attire, thereby improving robustness and fostering more inclusive AI for fashion design.

\section{Conclusion}\label{sec:conclusion}
In this paper, we introduced GarmentSketch, a large-scale dataset of more than 26,000 paired fashion sketches and detailed textual descriptions, created to advance multimodal image synthesis. Through comprehensive benchmarking of state-of-the-art generative models, we identified a central trade-off between photorealism and structural fidelity: MLLM approaches such as Gemini Nano Banana excel in visual quality, while generative models like ControlNet and diffusion models achieve superior adherence to sketch structure. GarmentSketch offers a valuable foundation for addressing these limitations and fostering research at the intersection of fashion design and generative modeling.

{Looking ahead, we plan to expand GarmentSketch to address current category imbalances and incorporate more culturally diverse attire. Furthermore, our future work will feature downstream fine-tuning experiments to adapt foundation models for domain-specific fashion synthesis. Finally, we will engage professional designers in human-centric evaluations to rigorously assess the quality and practical usability of these generated designs within real-world workflows.}

\section*{Acknowledgments}

This research is funded by Vietnam National University - Ho Chi Minh City (VNU-HCM) under Grant Number B2026-18-17.

\bibliographystyle{splncs04}
\bibliography{sample-base}

@inproceedings{nguyen-truc2025advancing,
  author    = {Nguyen-Truc, Nhu-Binh and Hoang, Nhu-Vinh and Nguyen, Tam V. and Tran, Minh-Triet and Le, Trung-Nghia},
  title     = {Advancing Fashion Design Through Intelligent Sketchpad Studio},
  booktitle = {Proceedings of the ACM International Conference on Multimedia},
  series    = {ACM MM '25},
  year      = {2025},
  @publisher = {Association for Computing Machinery},
}

@inproceedings{polyvore,
  author = {Han, Xintong and Wu, Zuxuan and Jiang, Yu-Gang and Davis, Larry S},
  title = {Learning Fashion Compatibility with Bidirectional LSTMs},
  booktitle = {ACM Multimedia},
  year  = {2017},
}

@misc{Fashion_product_images_dataset,
	title={Fashion Product Images Dataset},
	url={https://www.kaggle.com/ds/139630},
	DOI={10.34740/KAGGLE/DS/139630},
	publisher={Kaggle},
	author={Param Aggarwal},
	year={2019}
}

@InProceedings{ControlNetSD1.5,
  title={Adding Conditional Control to Text-to-Image Diffusion Models}, 
  author={Lvmin Zhang and Anyi Rao and Maneesh Agrawala},
  booktitle={ICCV},
  year={2023}
}

@InProceedings{SD,
        author    = {Rombach, Robin and Blattmann, Andreas and Lorenz, Dominik and Esser, Patrick and Ommer, Bj\"orn},
        title     = {High-Resolution Image Synthesis With Latent Diffusion Models},
        booktitle = {CVPR},
        month     = {June},
        year      = {2022},
        pages     = {10684-10695}
    }

@misc{SDXL,
      title={SDXL: Improving Latent Diffusion Models for High-Resolution Image Synthesis}, 
      author={Dustin Podell and Zion English and Kyle Lacey and Andreas Blattmann and Tim Dockhorn and Jonas Müller and Joe Penna and Robin Rombach},
      year={2023},
      eprint={2307.01952},
      archivePrefix={arXiv},
      primaryClass={cs.CV},
      url={https://arxiv.org/abs/2307.01952}, 
}

@article{t2i-adapter,
  title={T2i-adapter: Learning adapters to dig out more controllable ability for text-to-image diffusion models},
  author={Mou, Chong and Wang, Xintao and Xie, Liangbin and Wu, Yanze and Zhang, Jian and Qi, Zhongang and Shan, Ying and Qie, Xiaohu},
  journal={arXiv preprint arXiv:2302.08453},
  year={2023}
}

@article{pix2pix,
  title={Image-to-Image Translation with Conditional Adversarial Networks},
  author={Isola, Phillip and Zhu, Jun-Yan and Zhou, Tinghui and Efros, Alexei A},
  journal={CVPR},
  year={2017}
}

@inproceedings{CycleGAN2017,
  title={Unpaired Image-to-Image Translation using Cycle-Consistent Adversarial Networks},
  author={Zhu, Jun-Yan and Park, Taesung and Isola, Phillip and Efros, Alexei A},
  booktitle={ICCV},
  year={2017}
}

@inproceedings{VITON-HD,
  title={VITON-HD: High-Resolution Virtual Try-On via Misalignment-Aware Normalization},
  author={Choi, Seunghwan and Park, Sunghyun and Lee, Minsoo and Choo, Jaegul},
  booktitle={Proc. of the IEEE conference on computer vision and pattern recognition (CVPR)},
  year={2021}
}

@article{InforDraw,
	      title={Learning to generate line drawings that convey geometry and semantics},
	      author={Chan, Caroline and Durand, Fredo and Isola, Phillip},
	      booktitle={CVPR},
	      year={2022}
	      }

@misc{clip,
      title={Learning Transferable Visual Models From Natural Language Supervision}, 
      author={Alec Radford and Jong Wook Kim and Chris Hallacy and Aditya Ramesh and Gabriel Goh and Sandhini Agarwal and Girish Sastry and Amanda Askell and Pamela Mishkin and Jack Clark and Gretchen Krueger and Ilya Sutskever},
      year={2021},
      eprint={2103.00020},
      archivePrefix={arXiv},
      primaryClass={cs.CV},
      url={https://arxiv.org/abs/2103.00020}, 
}

@inproceedings{pictureThatSketch,
title={{Picture that Sketch: Photorealistic Image Generation from Abstract Sketches}},
author={Subhadeep Koley and Ayan Kumar Bhunia and Aneeshan Sain and Pinaki Nath Chowdhury and Tao Xiang and Yi-Zhe Song},
booktitle={CVPR},
year={2023}
}

@inproceedings{partialDiscrete,
author = {Sharma, Nakul and Tripathi, Aditay and Chakraborty, Anirban and Mishra, Anand},
year = {2024},
month = {06},
pages = {6024-6034},
title = {Sketch-guided Image Inpainting with Partial Discrete Diffusion Process},
doi = {10.1109/CVPRW63382.2024.00609}
}

@misc{fid,
      title={A Style-Based Generator Architecture for Generative Adversarial Networks}, 
      author={Tero Karras and Samuli Laine and Timo Aila},
      year={2019},
      eprint={1812.04948},
      archivePrefix={arXiv},
      primaryClass={cs.NE},
      url={https://arxiv.org/abs/1812.04948}, 
}

@inproceedings{lpips,
  title={The Unreasonable Effectiveness of Deep Features as a Perceptual Metric},
  author={Zhang, Richard and Isola, Phillip and Efros, Alexei A and Shechtman, Eli and Wang, Oliver},
  booktitle={CVPR},
  year={2018}
}

@misc{clipscore,
      title={CLIPScore: A Reference-free Evaluation Metric for Image Captioning}, 
      author={Jack Hessel and Ari Holtzman and Maxwell Forbes and Ronan Le Bras and Yejin Choi},
      year={2022},
      eprint={2104.08718},
      archivePrefix={arXiv},
      primaryClass={cs.CV},
      url={https://arxiv.org/abs/2104.08718}, 
}

@inproceedings{DeepFashion,
 author = {Liu, Ziwei and Luo, Ping and Qiu, Shi and Wang, Xiaogang and Tang, Xiaoou},
 title = {DeepFashion: Powering Robust Clothes Recognition and Retrieval with Rich Annotations},
 booktitle = {Proceedings of IEEE Conference on Computer Vision and Pattern Recognition (CVPR)},
 month = {June},
 year = {2016}
 }

@inproceedings{fashionAI,
  title={Fashionai: A hierarchical dataset for fashion understanding},
  author={Zou, Xingxing and Kong, Xiangheng and Wong, Waikeung and Wang, Congde and Liu, Yuguang and Cao, Yang},
  booktitle={CVPR Workshops},
  pages={0--0},
  year={2019}
}

@inproceedings{modanet,
  author       = {Shuai Zheng and Fan Yang and M. Hadi Kiapour and Robinson Piramuthu},
  title        = {ModaNet: A Large-Scale Street Fashion Dataset with Polygon Annotations},
  booktitle    = {ACM Multimedia},
  year         = {2018},
}

@inproceedings{viton,
  title = {VITON: An Image-based Virtual Try-on Network},
  author = {Han, Xintong and Wu, Zuxuan and Wu, Zhe and Yu, Ruichi and Davis, Larry S},
  booktitle = {CVPR},
  year  = {2018},
}

@article{sketchy,
  title={The sketchy database: learning to retrieve badly drawn bunnies},
  author={Sangkloy, Patsorn and Burnell, Nathan and Ham, Cusuh and Hays, James},
  journal={ACM Transactions on Graphics (TOG)},
  volume={35},
  number={4},
  pages={1--12},
  year={2016},
  publisher={ACM New York, NY, USA}
}

@article{quickdraw,
  author    = {David Ha and
               Douglas Eck},
  title     = {A Neural Representation of Sketch Drawings},
  journal   = {CoRR},
  volume    = {abs/1704.03477},
  year      = {2017},
  url       = {http://arxiv.org/abs/1704.03477},
  archivePrefix = {arXiv},
  eprint    = {1704.03477},
  bibsource = {dblp computer science bibliography, https://dblp.org}
}

@inproceedings{self-conditional-GAN,
  title={Self-supervised sketch-to-image synthesis},
  author={Liu, Bingchen and Zhu, Yizhe and Song, Kunpeng and Elgammal, Ahmed},
  booktitle={Proceedings of the AAAI conference on artificial intelligence},
  volume={35},
  number={3},
  pages={2073--2081},
  year={2021}
}

@inproceedings{quality-conditonal-GAN,
  title={Quality guided sketch-to-photo image synthesis},
  author={Osahor, Uche and Kazemi, Hadi and Dabouei, Ali and Nasrabadi, Nasser},
  booktitle={CVPR Workshops},
  pages={820--821},
  year={2020}
}

@inproceedings{democratisingsketch,
  title={It's All About Your Sketch: Democratising Sketch Control in Diffusion Models},
  author={Koley, Subhadeep and Bhunia, Ayan Kumar and Sekhri, Deeptanshu and Sain, Aneeshan and Chowdhury, Pinaki Nath and Xiang, Tao and Song, Yi-Zhe},
  booktitle={Proceedings of the IEEE/CVF Conference on Computer Vision and Pattern Recognition},
  pages={7204--7214},
  year={2024}
}

@article{wang2025jco,
  title={JCo-MVTON: Jointly Controllable Multi-Modal Diffusion Transformer for Mask-Free Virtual Try-on},
  author={Wang, Aowen and Li, Wei and Luo, Hao and Ao, Mengxing and Zhu, Chenyu and Li, Xinyang and Wang, Fan},
  journal={arXiv preprint arXiv:2508.17614},
  year={2025}
}

@article{shen2024IMAGDressing-v1,
    title={IMAGDressing-v1: Customizable Virtual Dressing},
    author={Shen, Fei and Jiang, Xin and He, Xin and Ye, Hu and Wang, Cong and Du, Xiaoyua and Zechao, Li and Tang, Jinhui},
    booktitle={arXiv preprint arXiv:2407.12705},
    year={2024}
  }

@inproceedings{jin2024human,
  title={Human-AI co-creation in fashion design ideation and sketching: an empirical study},
  author={Jin, Yu and Lee, Kyungho},
  booktitle={Proceedings of IEEE/CVF Computer Vision and Pattern Recognition Conference (CVPR), CVFAD Workshop, Seattle, USA},
  year={2024}
}

@misc{clipasso,
      title={CLIPasso: Semantically-Aware Object Sketching}, 
      author={Yael Vinker and Ehsan Pajouheshgar and Jessica Y. Bo and Roman Christian Bachmann and Amit Haim Bermano and Daniel Cohen-Or and Amir Zamir and Ariel Shamir},
      year={2022},
      eprint={2202.05822},
      archivePrefix={arXiv},
      primaryClass={cs.GR}
}

@misc{swiftsketch,
      title={SwiftSketch: A Diffusion Model for Image-to-Vector Sketch Generation}, 
      author={Ellie Arar and Yarden Frenkel and Daniel Cohen-Or and Ariel Shamir and Yael Vinker},
      year={2025},
      eprint={2502.08642},
      archivePrefix={arXiv},
      primaryClass={cs.CV},
      url={https://arxiv.org/abs/2502.08642}, 
}

@inproceedings{NguyenNgoc2023DMVTON,
  title = {DM-VTON: Distilled Mobile Real-time Virtual Try-On},
  author = {Nguyen-Ngoc, Khoi-Nguyen and Phan-Nguyen, Thanh-Tung and Le, Khanh-Duy and Nguyen, Tam V. and Tran, Minh-Triet and Le, Trung-Nghia},
  booktitle = {ISMAR},
  year = {2023},
}
\end{document}